\documentclass[letterpaper, 10 pt, conference]{ieeeconf}

\IEEEoverridecommandlockouts
\usepackage{glossaries}
\usepackage{amssymb}
\usepackage[normalem]{ulem}
\usepackage[usenames,dvipsnames,svgnames]{xcolor}
\usepackage{mathtools}
\usepackage{algorithmic}
\usepackage{algorithm}
\usepackage{dsfont}
\usepackage{subcaption}
\usepackage{mathrsfs}
\usepackage{ulem}

\newacronym{lsatc}{LSatC}{Low Earth Orbit Satellite Constellation}
\newacronym{fl}{FL}{Federated Learning}
\newacronym{nerf}{NeRF}{Neural Radiance Fields}
\newacronym{eo}{EO}{Earth Observation}
\newacronym{gnn}{GNN}{Graph Neural Network}
\newacronym{gat}{GAT}{Graph Attention Network}
\newacronym{magat}{MA-GAT}{Multi-Agent Graph Attention Network}
\newacronym{gatmarl}{GAT-MARL}{Graph Attention-based Multi-Agent Reinforcement Learning}
\newacronym{dtn}{DTN}{Delay-Tolerant Network}
\newacronym{ldtn}{LDTN}{Lunar Delay-Tolerant Network}
\newacronym{madrl}{MA-DRL}{Multi-Agent Deep Reinforcement Learning}
\newacronym{isl}{ISL}{Inter-Satellite Link}
\newacronym{nn}{NN}{Neural Network}
\newacronym{cgr}{CGR}{Contact Graph Routing}
\newacronym{iid}{IID}{Independent and Identically Distributed}
\newacronym{cadre}{CADRE}{Cooperative Autonomous Distributed Robotic Exploration}
\newacronym{r2l}{$\mathfrak{r}$2$\mathcal{L}$}{Rover-to-Lander}
\newacronym{r2r}{$\mathfrak{r}$2$\mathfrak{r}$}{Rover-to-Rover}
\newacronym{jpl}{JPL}{NASA's Jet Propulsion Laboratory}
\newacronym{gpr}{GPR}{Ground Penetrating Radars}
\newacronym{los}{LOS}{Line Of Sight}
\newacronym{rl}{RL}{Reinforcement Learning}
\newacronym{flop}{FLOP}{Floating Point Operations}
\newacronym{trl}{TRL}{Technology Readiness Level}
\newacronym{mimo}{MIMO}{Multiple-Input Multiple-Output}
\newacronym{e2e}{E2E}{End-to-End}
\newacronym{ml}{ML}{Machine Learning}
\newacronym{pomdp}{POMDP}{Partially Observable Markov Decision Problem}
\newacronym{fifo}{FIFO}{First-In, First-Out}
\newacronym{ctde}{CTDE}{Centralized Training, Decentralized Execution}
\newacronym{sgd}{SGD}{Stochastic Gradient Descent}
\newacronym{fp}{FP}{Frontier Point}
\newacronym{ttl}{TTL}{Time-To-Lander}
\newacronym{td}{TD}{Temporal Difference}
\newacronym{ddqn}{DDQN}{Double Deep Q-Learning}

\newif\ifmargincomments
\margincommentstrue

\ifmargincomments
\newcommand{\flc}[1]{\textcolor{teal}{#1}}

\newcommand{\asc}[1]{\textcolor{purple}{#1}}

\newcommand{\bs}[1]{\textcolor{blue}{#1}}

\newcommand{\frinline}[1]{\textcolor{RoyalBlue}{{\small \emph{#1}}}}

\else

\newcommand{\flc}[1]{{}}

\newcommand{\asc}[1]{{}}

\newcommand{\bs}[1]{{}}

\newcommand{\frinline}[1]{{}}

\fi

\title{\LARGE \bf
Learning Decentralized Routing Policies via Graph Attention-based Multi-Agent Reinforcement Learning in Lunar Delay-Tolerant Networks
}

\author{Federico Lozano-Cuadra$^{*}$, Beatriz Soret$^{*}$, Marc Sanchez Net$^{\dagger}$, Abhishek Cauligi$^{\ddagger}$, and Federico Rossi$^{\dagger}$%
\thanks{$^{*}$Federico Lozano-Cuadra and Beatriz Soret are with the Telecommunications Research Institute, University of Málaga, Spain.
{\tt\small flozano@ic.uma.es}}%
\thanks{$^{\dagger}$Marc Sanchez Net and Federico Rossi are with the Jet Propulsion Laboratory, California Institute of Technology, Pasadena, CA USA.}%
\thanks{$^{\ddagger}$Abhishek Cauligi is with the Department of Mechanical Engineering, Johns Hopkins University, Baltimore, MD USA.}%
\thanks{The work of F. Lozano-Cuadra and B. Soret is partially supported by the Spanish Ministerio de Ciencia e Innovación under grant PID2022-136269OB-I00 funded by MCIN/AEI/10.13039/501100011033 and “ERDF A way of making Europe”. Part of this research was carried out at the Jet Propulsion Laboratory, California Institute of Technology, under a contract with the National Aeronautics and Space Administration (80NM0018D0004).}
}

\begin{document}

\maketitle
\begin{abstract}
We present a fully decentralized routing framework for multi-robot exploration missions operating under the constraints of a \gls{ldtn}. In this setting, autonomous rovers must relay collected data to a lander under intermittent connectivity and unknown mobility patterns. We formulate the problem as a \gls{pomdp} and propose a \gls{gatmarl} policy that performs \gls{ctde}. Our method relies only on local observations and does not require global topology updates or packet replication, unlike classical approaches such as shortest path and controlled flooding-based algorithms. Through 
Monte Carlo simulations in randomized exploration environments, \gls{gatmarl} provides higher delivery rates, no duplications, and fewer packet losses, and is able to leverage short-term mobility forecasts; offering a scalable solution for future space robotic systems for planetary exploration, as demonstrated by successful generalization to larger rover teams. 
\end{abstract}
\vspace{-.1cm}


\glsresetall

\section{Introduction}

\noindent The renewed interest in planetary and lunar surface exploration has accelerated the development of autonomous multi-robot systems. 
Missions such as \gls{cadre}~\cite{rabideau2025planning}, led by \gls{jpl}, aim to demonstrate high \gls{trl} for cooperative autonomy in space robotics. In particular, \gls{cadre} plans to deploy a team of three rovers to explore the Reiner Gamma region of the Moon around a lander, which will serve as a base station and gateway to Earth. 

\noindent However, these initial demonstrations assume that the exploring agents remain within the communication range of the lander, thereby simplifying network design because nodes (rovers and lander) can simply broadcast the information {and assume all-to-all communication}. In future science-driven missions, exploring agents will need to leave the lander's coverage area to reach farther, uncharted regions. This shift introduces new challenges: nodes must operate autonomously as a \gls{dtn}~\cite{fall2003delay}, making independent routing and storage decisions under intermittent connectivity. {This task is especially challenging in an unknown and complex environment like the lunar surface, further complicating the problem of reliable transmission of data packets and effective network management.} 
As such, there is a pressing need to develop new routing algorithms  for the next generation of autonomous networks and distributed robotic explorers. 
In this paper, we address these challenges by proposing a decentralized, learning-based communication policy where each rover makes local routing decisions based on limited neighborhood observations (Fig.~\ref{fig:LDTN}), enabling robust multi-agent collaboration under \gls{dtn} conditions.

\noindent The problem of routing in time-varying networks has seen significant interest from the community: 


\begin{figure}[t]
    \centering
    {\includegraphics[width=0.48\textwidth]{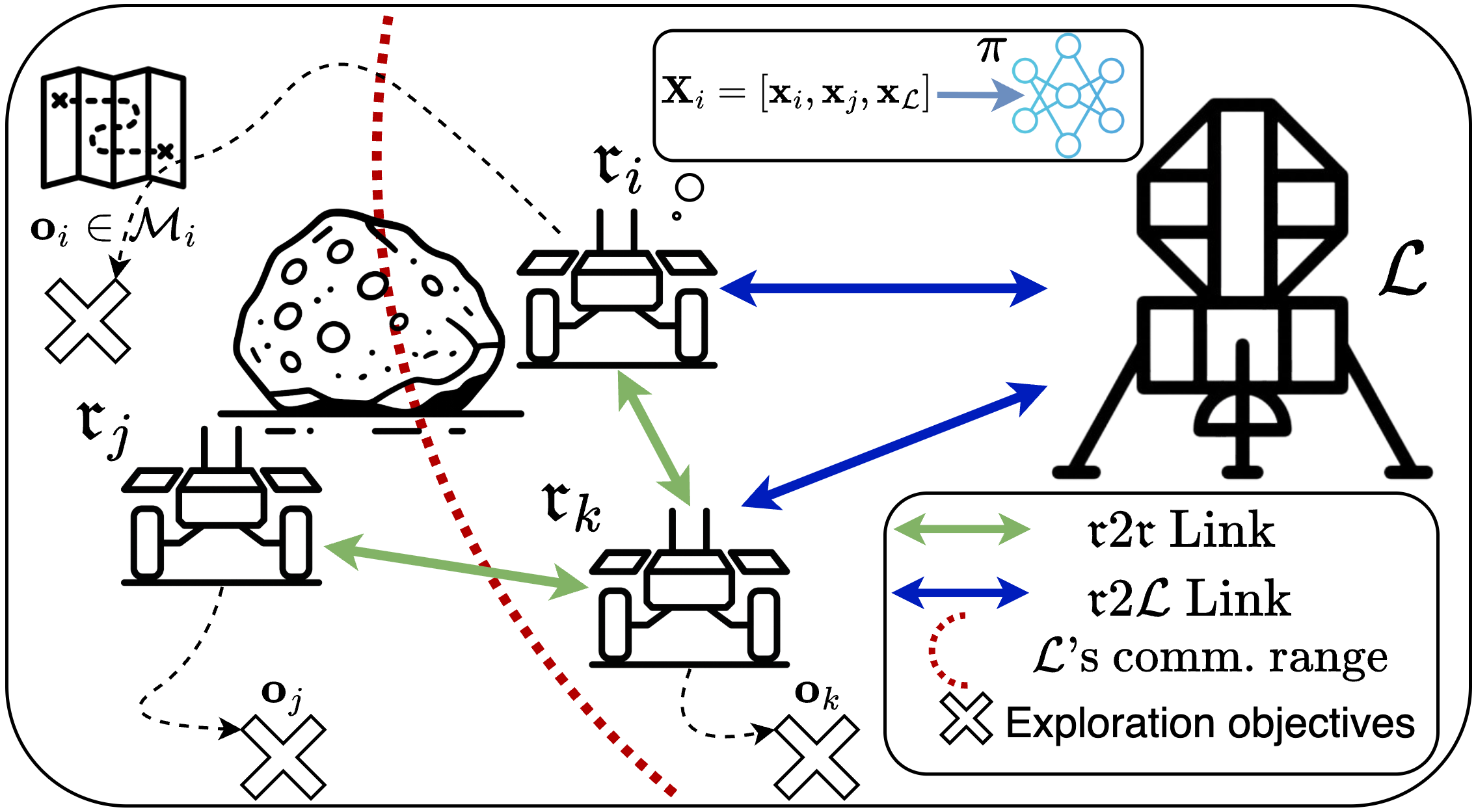}}
    \caption{Snapshot of a Lunar Delay Tolerant Network (LDTN) with three rovers $\mathfrak{r}\in\mathfrak{R}$ and a lander $\mathcal{L}$, which serves as the base station.
    Rovers $\mathfrak{r}_i$ and $\mathfrak{r}_k$ have direct connection with $\mathcal{L}$ and with one another, whereas $\mathfrak{r}_j$ can only communicate with $\mathfrak{r}_k$. 
    Every rover explores a different sub-area of the map $\mathcal{M}$ by defining objectives $\mathbf{o}$.
    Each $\mathfrak{r}$ collects a local observation $\mathbf{X}_r = [\mathbf{x}_\mathfrak{n}]_{\mathfrak{n} \in \mathcal{N}_r}$ from its neighborhood (e.g., $\mathcal{N}_i = \{\mathfrak{r}_i, \mathfrak{r}_k, \mathcal{L}\}$ for $\mathfrak{r}_i$), which serves as input to a pre-trained policy $\pi$ that determines the routing decisions. 
    }
    \label{fig:LDTN}\vspace{-0.5cm}
\end{figure}


\noindent {\emph{Delay Tolerant Networks:} }
\glspl{dtn} are communication networks designed for environments with long propagation delays and intermittent or disrupted connectivity. 
Their architectures introduce store-and-forward mechanisms, enabling data packets to be buffered at the source or at intermediate nodes until a suitable communication opportunity arises~\cite{fall2003delay}. This paradigm is particularly relevant for the networks {found in} interplanetary exploration~\cite{burleigh2003delay}, where continuous \gls{e2e} paths rarely exist and  link availability is highly dynamic. 
The extensive literature on the topic of routing in \glspl{dtn} can be classified according to the available \emph{information oracles}~\cite{jain2004routing}. The first category comprises policies that rely solely on local knowledge of 1-hop neighbors. Here, random algorithms such as \emph{Hot Potato}~\cite{hotpotato},  controlled flooding algorithms like \emph{Spray and Wait}~\cite{spyropoulos2005spray}, and opportunistic routing~\cite{net2018evaluation} are applicable. A second category includes algorithms that require knowledge of the network topology and link availability. Classical shortest path algorithms like \emph{Dijkstra}~\cite{dijkstra1959} belong to this category, potentially including queueing information if available. 
Another family of policies depends on knowledge of node mobility. For example, \emph{\gls{cgr}} predicts \gls{e2e} paths for packets using pre-computed contact opportunities between all nodes in the network~\cite{araniti2015contact}. 
Finally, genie-aided oracles have full knowledge of the system, including current and future traffic demands {in addition to future contact opportunities.}
{In such cases, linear programming} provides the optimal solution~\cite{jain2004routing}. 
However, in autonomous robotic exploration scenarios, access to such information oracles is typically {infeasible} 
due to intermittent connectivity, unknown mobility patterns, and the need for decentralized decision-making, which arises naturally from these properties of the communication network. 
While traditional \gls{dtn} routing methods perform well under specific assumptions, they face limitations in scenarios with partial observability, decentralized operation, or highly dynamic topologies. Data-driven methods can provide additional performance by learning patterns in traffic demand and contact availability, improving performance even when explicit models are unavailable or inaccurate.




\noindent\emph{Machine Learning for distributed space systems:} 
\gls{ml} techniques are increasingly being explored for autonomous decision-making in space communication systems. 
For planetary exploration scenarios,~\cite{SzatmariCauligi2025} studies the use of \gls{fl} to merge the maps generated by exploring rovers in a mission concept inspired by \gls{cadre}'s concept of operations.
In~\cite{Harkavy2020}, a \gls{rl} algorithm is proposed to autonomously manage the buffer of one node in a \gls{dtn}. 
In~\cite{10624807, lozanocuadra2024continualdeepreinforcementlearning} multi-agent \gls{rl} is proposed to solve the routing problem in satellite constellations, where each node of the network is an independent learning agent that makes its own autonomous routing decisions. 
However, the works in~\cite{10624807, lozanocuadra2024continualdeepreinforcementlearning} assume continuous \gls{e2e} connectivity and do not incorporate store-and-forward mechanisms, {rendering} them unsuitable for \glspl{dtn}.
Additionally, their models do not support variable-sized input graphs, limiting their adaptability in dynamic multi-robot settings {in which rovers may split and regroup as they explore.} 
As a result, direct comparison with existing \gls{ml}-based routing methods is not straightforward, since most rely on continuous connectivity or centralized knowledge. 
In contrast, \glspl{gat}~\cite{velivckovic2017graph} naturally accommodate changing neighborhood sizes, making them well suited for decentralized cooperation in evolving topologies. 
Building on graph-based methods,~\cite{DolanNayakEtAL2025} proposes a multi-agent \gls{rl} approach that processes graph-structured state information to enable decentralized satellite coordination under limited communication. 






\noindent\emph{Statement of Contributions:} 
\noindent The main contributions of this paper are: 
(1) we {propose a rigorous decision-theoretic model of} 
routing over a \gls{ldtn} in the context of planetary exploration missions, 
applicable to any beyond-line-of-sight robotic networks; 
(2) we 
propose a novel {\gls{gatmarl}}
policy that performs decentralized routing under intermittent connectivity and partial observability;
(3) we demonstrate through extensive simulations in a custom Python-based packet-level simulator that \gls{gatmarl} outperforms classical \gls{dtn} strategies, including \emph{Spray and Wait} and \emph{Greedy Forwarding}, while demonstrating scalability and relying only on locally observed information.


\noindent\emph{Organization:} The rest of this paper is organized as follows. In Section~\ref{sec:systemmodel}, we present a rigorous model of the proposed~\gls{ldtn} and of its concept of operations. Section~\ref{sec:P} formalizes the packet routing problem.  Section~\ref{sec:Learning_framework} introduces the proposed \gls{gatmarl} framework and Section~\ref{sec:results} explores its performance through extensive simulations. Finally, we conclude in Section~\ref{sec:conclusions}.



\section{System Model} \label{sec:systemmodel}

\noindent Figure~\ref{fig:LDTN} shows the considered \gls{ldtn} with three primary components: (1) A {team} 
of $\mathfrak{R}$ exploring lunar rovers; (2) a lander $\mathcal{L}$, serving as a base station; and (3) the map region of the lunar surface to be explored $\mathcal{M}$, characterized by obstacles such as craters and rocks that hinder rover mobility and communications. 
This system model is analogous to \gls{cadre}'s~\cite{rabideau2025planning}, a planned mission which involves the deployment of three cooperative robotic explorers to the lunar surface. 
Although the rovers are moving and the related communication parameters change with time, we omit the time index for notation simplicity. 

\noindent Each rover $\mathfrak{r} \in \mathfrak{R}$ is a wheeled vehicle that explores $\mathcal{M}$ according to specific exploration objectives. During this exercise, each $\mathfrak{r}$ gathers information and encapsulates it in data packets $p_\mathfrak{r} \in \mathcal{P}_\mathfrak{r}$, with the global set of packets defined as $\mathcal{P} = \bigcup_{\mathfrak{r}\in \mathfrak{R}} \mathcal{P}_\mathfrak{r}$. Each rover $\mathfrak{r}$ is able to temporarily store packets in its local buffer $\mathcal{B}_\mathfrak{r}$ of capacity $B$, thereby forming a \gls{ldtn} (Fig.~\ref{fig:LDTN}). 
The joint objective of the rover {team} 
$\mathfrak{R}$ is to maximize the number of packets $p \in \mathcal{P}$ that are transmitted to lander $\mathcal{L}$ through the \gls{ldtn} without being dropped or lost. 
$\mathcal{L}$ is modeled as an infinite-capacity sink node for two main reasons: (1) {it} is equipped with a significantly larger buffer, $\mathcal{B}_{\mathcal{L}}$, and (2) it serves as a gateway to Earth, maintaining a direct, {high-bandwidth (albeit intermittent)} communication link that allows it to offload data as needed.


\noindent Each $\mathfrak{r}$ is equipped with a mesh radio module (e.g., the Microhard pMDDL used in CADRE) 
which supports \gls{mimo} operation and enables the establishment of direct communication links with the lander $\mathcal{L}$, referred to as \gls{r2l} links, as well as with other rovers, forming \gls{r2r} links. 
Moreover, rovers can also do store-and-forward (where $\mathfrak{r}$ temporarily keeps the packets in its $\mathcal{B}_\mathfrak{r}$ until a link to the intended destination node is available). 
Mathematically, the network is modeled as a time-varying graph $\mathcal{G} = (\mathcal{N}, \mathcal{E})$, where $\mathcal{N} = \{ \mathfrak{R} \cup \mathcal{L}\}$ is the set of nodes, and $\mathcal{E} = \bigcup_{\mathfrak{r}\in \mathfrak{R}} \mathcal{E}_\mathfrak{r}$ is the union of all dynamically established communication links/edges. 
$\mathcal{E}_\mathfrak{r}$ 
{is the set of potential edges}
for each $\mathfrak{r}$, with $|\mathcal{E}_\mathfrak{r}| \leq |\mathfrak{R}| + 1$, as it can connect with any other $\mathfrak{r} \in \mathfrak{R}$ (including themselves) and with $\mathcal{L}$, and the local neighborhood $\mathcal{N}_\mathfrak{r}$ is the set containing $\mathfrak{r}$ and all nodes it shares a communication edge with. 
Together, these edges enable the formation of a dynamic wireless mesh network~\cite{saboia2024cadre} over the lunar surface. 

\noindent \textbf{Exploration area: }
$\mathcal{M}$ denotes the region of the lunar surface surrounding the lander that is assigned for rover exploration. This area consists of traversable terrain interspersed with obstacles such as craters and rocks that hinder rover mobility and communications. The spatial distribution and sizing of obstacles within $\mathcal{M}$ are representative and determined using a probabilistic model calibrated with empirical lunar surface data provided by the \gls{cadre} mission.  

\begin{figure}[t]
    \centering
    {\includegraphics[width=0.48\textwidth]{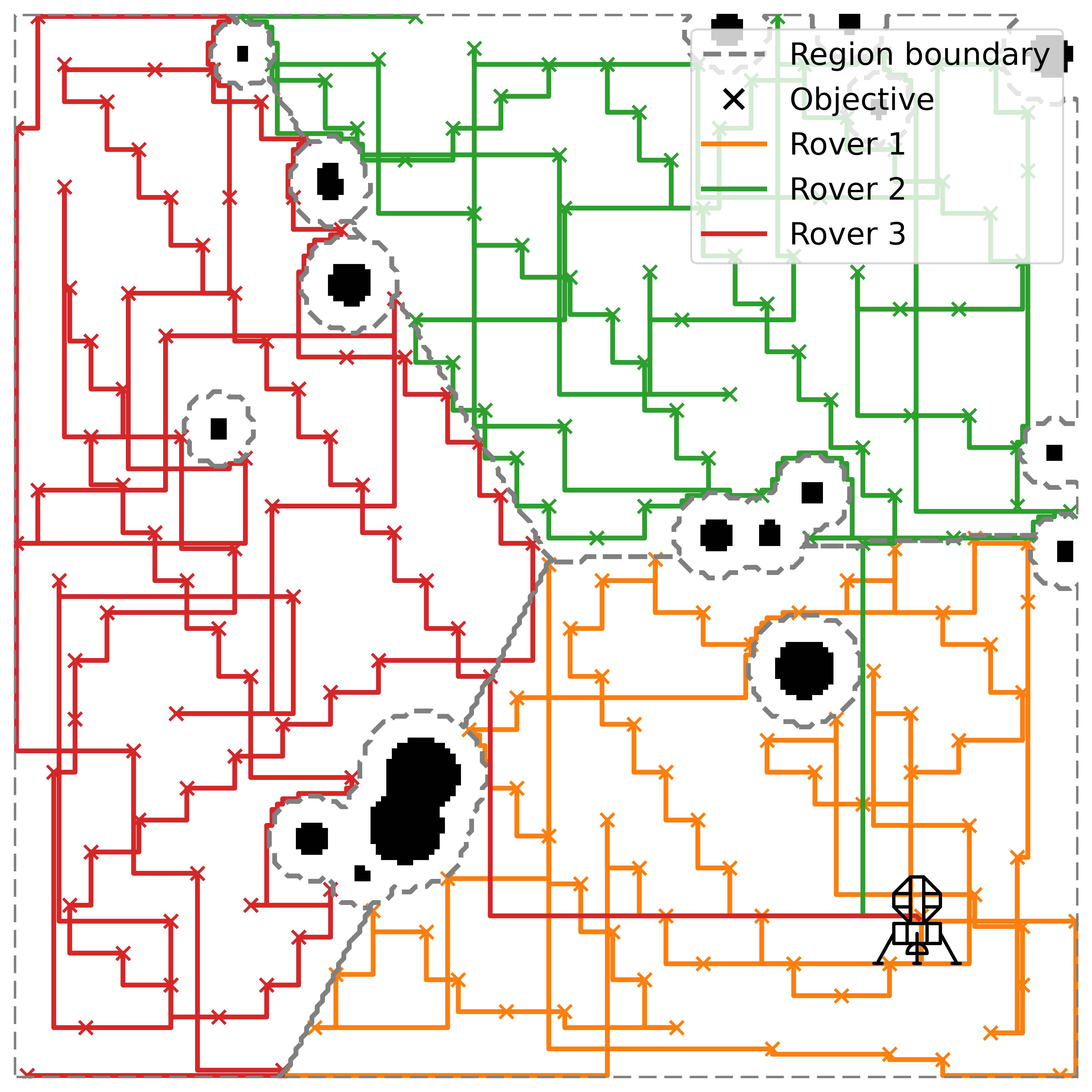}}
    \caption{The exploration region $\mathcal{M}$ is partitioned into sub-regions, each assigned to a different rover. Rovers depart from the lander and autonomously explore their assigned areas, selecting frontier points as navigation objectives while avoiding obstacles.}
    \label{fig:Exploration}\vspace{-0.7cm}
\end{figure}

\noindent \textbf{Rover exploration:} 
{The rovers explore according to the multi-agent exploration algorithm developed for the CADRE mission. The algorithm is briefly described here for convenience and we refer the interested reader to~\cite{nayak2025multi} for a detailed description.}
An exploration episode $\tau$ starts with every $\mathfrak{r} \in \mathfrak{R}$ at $\mathcal{L}$'s position (Fig.~\ref{fig:Exploration}).  
At the beginning of $\tau$, the boundary of the region to be explored $\mathcal{M}$ is defined 
{(in \gls{cadre}, this is specified by ground operators~\cite{rabideau2025planning})}. 
$\mathcal{M}$ is first partitioned into $|\mathfrak{R}|$ sub-regions using \emph{K-Means} clustering (Fig.~\ref{fig:Exploration}), and each sub-region $\mathcal{M}_\mathfrak{r} \subseteq \mathcal{M}$ is then assigned to a rover $\mathfrak{r} \in \mathfrak{R}$ using the \emph{Hungarian algorithm}~\cite{nayak2025multi} to minimize the total expected travel cost. 
$\mathcal{M}$ is discretized into $\mathfrak{m}$ square cells of resolution $\rho$. A discrete time step $t$ corresponds to the time it takes a rover to move from one cell to an adjacent cell. 
The content of $\mathcal{M}$ is initially unknown to the rovers, and their collective exploration objective is to observe and classify every reachable portion of $\mathcal{M}$ as “traversable” or “obstacle” using on-board stereo cameras. Each $\mathfrak{r}$ autonomously explores its assigned $\mathcal{M}_\mathfrak{r}$ using frontier-based exploration{~\cite{yamauchi1997frontier, nayak2025multi}}, 
incrementally mapping and labeling the terrain. 
Every $\mathfrak{r}$ selects its next navigation objective among a set of frontier points independently based on a local observation, enabling decentralized and adaptive exploration throughout the region. All frontier points $f$ are evaluated using a weighted score:
%
\begin{equation*}
    \text{score}(f) = w_1 \cdot d(f) - w_2 \cdot n(f),
\end{equation*}
 where $d(f)$ is the Manhattan distance from $\mathfrak{r}$ to $f$, $n(f)$ is the information gain, i.e, the number of cells from $\mathcal{M}_\mathfrak{r}$ that would potentially be explored by $\mathfrak{r}$ along its path to $f$, and $w_1$, $w_2$ are tunable weights. 
To prevent rover collisions, obstacles are {inflated using a circular footprint} (Fig.~\ref{fig:Exploration}). The chosen $f$ becomes the next objective $\mathbf{o}$ for $\mathfrak{r}$, which computes the path towards it in advance. 
Frontier points are classified as unreachable if no collision-free path exists to reach them due to surrounding obstacles. 
The exploration episode $\tau$ ends when all the reachable points of $\mathcal{M}$ have been classified.

\noindent \textbf{Packet generation:} 
During exploration, rovers collect imagery required for photogrammetric 3D reconstruction 
of the surface without human intervention and subsequently perform distributed sensing. 
While navigating between objectives, each rover $\mathfrak{r}$ collects the information and
incrementally builds local maps of the surface. 
These maps classify regions as “traversable” or “obstacle” and support autonomous navigation and decision-making. 
Each $\mathfrak{r}$ periodically encapsulates this information in one packet $p_\mathfrak{r}(t)$ every $F$ time steps $t$ and stores it in its local buffer $\mathcal{B}_\mathfrak{r}$. 
Each packet corresponds to a 1456-byte compressed traversability map, matching the telemetry format used by \gls{cadre}’s mapping component. It includes timestamps, sequence identifiers, reference vectors, orientation data, and four resolution levels of compressed binary maps, and is generated at 1Hz. 


\noindent \textbf{Link establishment:} 
As the rovers move during exploration, the network topology evolves dynamically; with communication links forming or breaking over time. A communication link $e_{ij}$ between nodes $i$ and $j$ is established at time $t$ if two conditions are met: 
(1) The distance between nodes satisfies $d(i, j) < d^{\max}$, where $d^{\max}$ is the maximum communication distance—a fixed parameter calibrated during pre-flight testing on Earth. This value is set depending on whether $e_{ij}$ is \gls{r2r} or \gls{r2l}. 
(2) The path between $i$ and $j$ must be clear of obstructions, captured by $\text{clear}(i,j)$. Therefore:
\vspace{-0.1cm}
\begin{equation*}
    e_{ij} \in \mathcal{E} \iff d(i, j) < d^{\max} \wedge \text{clear}(i,j).
\vspace{-0.1cm}
\end{equation*}

\noindent If these conditions are satisfied with multiple nodes, a rover can maintain simultaneous active links to all such neighbors, each with its own data rate, thereby forming a dynamic mesh network. 
Fig.~\ref{fig:LDTN} illustrates the resulting topology, highlighting both available links and those that are unavailable due to either obstacles or distance limitations.







\noindent \textbf{Data rate:} 
The achievable data rate for each established link $e_{ij}$ is denoted by $R(i, j)$ and is defined as:

%
\begin{equation}
    R(i, j) =
    \begin{cases}
    R_\mathfrak{r}^{\max}, & \text{if } i, j \in \mathfrak{R} \ (\text{\gls{r2r} link}) \\
    R_{\mathcal{L}}^{\max}, & \text{if } \{i, j\} \ni \{\mathcal{L}\} \ (\text{\gls{r2l} link}) \\
    0, & \text{if } e_{ij} \notin \mathcal{E}
    \end{cases},
    \label{eq:Rate}
\end{equation}
%
\noindent where $R_\mathfrak{r}^{\max}$ and $R_{\mathcal{L}}^{\max}$ represent the maximum achievable data rates for \gls{r2r} and \gls{r2l} links, respectively. 
As $\mathcal{L}$'s antenna has a higher elevation, we note that $R_\mathcal{L}^{\max} > R_\mathfrak{r}^{\max}$. 
These values are inspired by Microhard pMDDL modules. Although the modules support adaptive rates, we assume constant rate links for simplicity and to focus on routing; the data rate does not vary with distance in our model.



\noindent \textbf{Routing:} 
For managing locally stored packets, each rover autonomously decides among the following actions for each packet: 
(1) hold packets in its buffer in order to perform store-and-forward, delaying transmission until connectivity conditions improve; (2) 
forward packets towards another node; 
or (3) generate and forward duplicate copies of packets through multiple links. 

\noindent \textbf{Queue Management:}
Each $\mathfrak{r}$ stores every generated packet $p_{\mathfrak{r}}$ in its local buffer $\mathcal{B}\mathfrak{r}$, which operates as a \gls{fifo} queue. When a new $p_{\mathfrak{r}}$ is created, it is appended to the end of $\mathcal{B}_\mathfrak{r}$. If $|\mathcal{B}_\mathfrak{r}|=B$ (full), the oldest packet is dropped to make room for a new one. At each $t$, $\mathfrak{r}_i$ may transmit up to $\min(R(i,j)\cdot t, |\mathcal{B}\mathfrak{r}|)$ packets to a selected neighbor $\mathfrak{r}_j$. Packets are forwarded in \gls{fifo} order, ensuring that the oldest packets are prioritized for transmission.



\section{Problem formulation} \label{sec:P}

\noindent The routing problem involves delivering {as many data packets $p$ as possible} from any rover $\mathfrak{r}$ to the lander $\mathcal{L}$ through the \gls{ldtn} with the limited available information on board each $\mathfrak{r}$ while satisfying all network constraints. 
Moreover, {we wish to} optimize
energy resources while minimizing the communication overhead. 
We formulate the problem as:
%
\begin{subequations}
  \begin{align}
                & {\mathbf{P}}: \max_{\mathbf{R}_{\mathfrak{r}\mathcal{L}}^\tau}\sum_{{p}=1}^{{|\mathcal{P}_\mathfrak{r}|}} \mathrm{D}^p(\mathfrak{r}_p,\mathcal{L}),  \quad &&{\forall \mathfrak{r} \in \mathfrak{R}}, \label{eq:problem}\\
    s.t. \quad  
                & f_{ij} \leq R(i, j), && \forall i, j \in \mathcal{N}, \label{eq:constraint_1}\\
                & |\mathcal{B}_\mathfrak{r}| \leq B, && \forall \mathfrak{r} \in \mathfrak{R}, \label{eq:constraint_3}
\end{align}
\label{eq:problem_formulation}
\vspace{-0.5cm}
\end{subequations}
%

\noindent {where $\mathrm{D}^p(\mathfrak{r}_p,\mathcal{L})$ is an indicator function that equals 1 if a unique copy of packet $p_\mathfrak{r}$ generated by rover $\mathfrak{r}$ is delivered to the lander $\mathcal{L}$, and 0 otherwise}. 
{$\mathbf{R}^\tau$ is the set of all possible routes or paths for routing data packets during an exploration episode $\tau$}, 
$\mathbf{R}_{\mathfrak{r}\mathcal{L}}^\tau \subseteq \mathbf{R}^\tau$ is 
{the subset of routes that start at $\mathfrak{r}$ and end at $\mathcal{L}$ to be optimized}, 
{and $f_{ij}$ is the total flow of packets that go from node $i$ to node $j$ at each time step $t$.} 
In the constraints, Eq.~\eqref{eq:constraint_1} states that $f_{ij}$ cannot exceed the capacity between the nodes, given by the achievable data rate $R(i,j)$ (Eq.~\eqref{eq:Rate}), limiting the usage of any given route. 
Eq.~\eqref{eq:constraint_3} imposes that the number of packets stored in the buffer of rover $\mathfrak{r}$, $|\mathcal{B}_\mathfrak{r}|$, does not exceed its capacity $B$. 
We note that no flow conservation constraints are imposed, since packets may be duplicated or dropped due to buffer overflows; similarly, no self-loop constraints are enforced, since rovers may perform store-and-forward.


\noindent \textbf{Discussion:} Solving Problem~\eqref{eq:problem_formulation} requires full knowledge of present and future link availability and packet flow—information that is not accessible in decentralized exploration missions. Furthermore, agents operate autonomously with only partial observations and no access to the states observed by others, making the problem a \gls{pomdp} from each agent’s perspective. We therefore address this decentralized, partially observable routing task using a \gls{rl} approach under the \gls{ctde} paradigm.

\section{Learning Framework} \label{sec:Learning_framework}
\noindent Each rover $\mathfrak{r} \in \mathfrak{R}$ operates as an independent agent, making routing decisions based solely on local observations in a partially observable, dynamic environment. The proposed \gls{gatmarl} framework follows a \gls{ctde} paradigm structured into two phases: (1) a centralized server aggregates the experiences from all rovers in a global buffer $D_g$ to train a shared policy model $\pi$ and (2) once trained, $\pi$ is deployed onboard each rover, enabling decentralized execution of routing decisions without further training. This design ensures that all rovers act in coordination under the same globally optimized policy. 
In flight operations, no further training is expected—only the trained policy $\pi$ is uploaded, and all routing decisions are made onboard fully decentralized.

\noindent We formulate the routing problem as a \gls{pomdp}, defined by the tuple $(\mathcal{S}, \mathcal{A}, P(s,a), \mathcal{R}(s,a))$, with discrete time step $t$. At each $t$, a rover $\mathfrak{r}_i$ observes a local state $S_t^i \in \mathcal{S}$, selects an action $a_t \in \mathcal{A}$, receives a reward $r_t$, 
{and transitions to a new state $S_{t+1}^j \in \mathcal{S}$ observed at the receiving node $\mathfrak{n}_j$ according to the transition probability $P(s_{t+1}^j \mid s_t^i, a_t)$.}

\noindent The interaction model follows the innovative \emph{packet-centric} multi-agent routing approach described in~\cite{10624807, lozanocuadra2024continualdeepreinforcementlearning}, where the state transition from $\mathfrak{r}_i$ to $\mathfrak{n}_j$ after forwarding packet $p$ is captured by the tuple: $(S_t^{i}, a_t^{i}, r_t^{i}, S_{t+1}^{j})$, where $S_t^{i}$ is observed at $\mathfrak{r}_i$, and $S_{t+1}^{j}$ is observed at $\mathfrak{r}_j$. Unlike~\cite{10624807,lozanocuadra2024continualdeepreinforcementlearning}, self-loops are allowed, i.e., $\mathfrak{r}_i$ may forward a packet to itself ($\mathfrak{r}_j = \mathfrak{r}_i$).
This tuple representation is particularly well suited for decentralized routing, as learning occurs from the perspective of the packet while the model remains fully local to each rover, resulting in an efficient and communication-light implementation. 

\noindent\textbf{State space:} 
At each decision step, a rover $\mathfrak{r}_i$ observes a local subgraph composed of itself and its 1-hop neighbor nodes $\mathfrak{n} \in \mathcal{N}_\mathfrak{r}$. 
For each $\mathfrak{n}$ in this subgraph, a {feature vector}
$\mathbf{x}_\mathfrak{n} \in \mathbb{R}^7$ is constructed:
%
\begin{equation*}
    \mathbf{x}_\mathfrak{n} = [\mathds{1}_{\{\mathfrak{n} = \mathfrak{r}\}}, \mathds{1}_{{\mathfrak{n} = \mathcal{L}}}, \mathds{1}_{{e_{\mathfrak{n}\mathcal{L}} \in \mathcal{E}}}, \text{TTL}_\mathfrak{n}, |\mathcal{B}_{\mathfrak{n}}|, d(\mathcal{M}_\mathfrak{n}, \mathcal{L}), d(\mathfrak{n}, \mathcal{L})], 
\end{equation*}
%
\noindent where each component is scaled in the range $[0, 10]$, $d(\mathfrak{n}, \mathcal{L})$ and $d(\mathcal{M}_\mathfrak{n}, \mathcal{L})$ are the Euclidean distances from $\mathfrak{n}$ and $\mathcal{M}_\mathfrak{n}$ to $\mathcal{L}$, respectively. 
Moreover, since rovers pre-compute their path towards their next $\mathbf{o}$, each $\mathfrak{n}$ can predict their \gls{ttl} value, $\text{TTL}_\mathfrak{n}$: the number of $t$ in which $\mathfrak{n}$ will establish a link $e_{\mathfrak{n}\mathcal{L}}$. 
These values form the node feature matrix $\mathbf{X}_\mathfrak{r} = [\mathbf{x}_\mathfrak{n}]_{\mathfrak{n} \in \mathcal{N}_\mathfrak{r}}$. Additionally, an adjacency matrix $\mathbf{A}_\mathfrak{r}$ encodes the bidirectional edges between the current node and each neighbor. 
A \gls{ttl}-masking mechanism creates an updated adjacency matrix $\mathbf{A}_\mathfrak{r}'$ that filters neighbors {with $\gls{ttl} \neq 0$}  
whenever they are available, i.e., nodes whose path will bring them to have a communication link with $\mathcal{L}$. 
Therefore, the observed state $S_t^i$ for agent $\mathfrak{r}_i$ is defined as the pair $(\mathbf{X}_\mathfrak{r}, \mathbf{A}_\mathfrak{r}')$, comprising the feature matrix of its local subgraph $\mathcal{N}_\mathfrak{r}$ and the corresponding masked adjacency matrix, respectively. 
Overall, this design ensures that the state space $\mathcal{S}$ remains fully local—no global topology or centralized knowledge is assumed. 

\noindent \textbf{Action space:} 
At each step, every rover $\mathfrak{r}$ selects an available edge $e_{ij} \in \mathcal{E}_\mathfrak{r}$—including self-loops—to transmit packets at rate $\min(R(i,j)\cdot t, |\mathcal{B}_i|)$. This captures both forwarding and storing actions. 

\noindent \textbf{Rewards:} 
The reward function is a composition of different terms that incentivize packet delivery while discouraging buffer congestion and inefficient forwarding from a rover $\mathfrak{r}_i$ its neighbor $\mathfrak{n}_j$. First, $r_{\text{TTL}}(\mathfrak{n}_j)$ is defined as follows:
%
\begin{equation}
    r_{\text{TTL}}(\mathfrak{n}_j) = 
    \begin{cases}
        r_{\text{noTTL}}, & \text{if } \text{TTL}_{\mathfrak{n}_j} = 0 \\
        r_{\text{TTL}} \cdot \frac{10^{\alpha_\text{TTL} \cdot \text{TTL}_{\mathfrak{n}_j}} - 1}{10^{\alpha_\text{TTL}} - 1} & \text{otherwise}
    \end{cases}.
    \label{eq:TTL}
\end{equation}
%
\noindent Then, the buffer usage penalty $r_{\text{B}}(\mathfrak{n}_j)$ is defined as:
%
\begin{equation}
    r_{\text{B}}(n_j) = r_{\text{usage}} \cdot \frac{10^{\alpha_B \cdot u_{n_j}} - 1}{10^{\alpha_B} - 1},
    \label{eq:buffer}
\end{equation}
%
\noindent where $u_{n_j} = |\mathcal{B}_{n_j}|/B$ is the normalized buffer usage at node $n_j$. 
In both Eqs.~\eqref{eq:TTL} and~\eqref{eq:buffer}, the reward functions follow a positive and negative exponential curve where $\alpha_\text{TTL}$ and $\alpha_B$ determine the steepness of each one, respectively. Therefore, the overall reward can be defined as:
%
%
\begin{equation}
    \begin{aligned}
        r =\ & r_{\text{B}}(n_j) + r_{\text{TTL}}(n_j) \\
        &+ \mathds{1}_{\{n_j = R_i\}} \cdot r_{\text{hold}} 
        + \mathds{1}_{\{n_j \neq R_i \land n_j \neq \mathcal{L}\}} \cdot r_{\text{fwd}} \\
        &+ \mathds{1}_{\{n_j = \mathcal{L}\}} \cdot r_{\text{deliver}} 
        + \mathds{1}_{\{\text{connected}_{n_j} = 1\}} \cdot r_{\text{conn}}
    \end{aligned},
    \label{eq:reward}
\end{equation}
%
\noindent 
where, $r_{\text{noTTL}}, r_{\text{B}}, r_{\text{hold}}, r_{\text{fwd}} \leq 0 \leq r_{\text{TTL}}, r_{\text{deliver}}, r_{\text{conn}}$. 
{$r_{\text{hold}}$ and $r_{\text{fwd}}$ are penalties for holding and forwarding packets, and $r_{\text{deliver}}$ and $r_{\text{conn}}$ are rewards for delivering packets to lander and to nodes that are connected to the lander, respectively.}

\noindent \textbf{Q-learning formulation:}
For {the proposed} decentralized routing setting, the Q-value update for rover $\mathfrak{r}_i$ selecting action $a_t^i$ at state $S_t^i$ is given by:
%
\begin{equation*}
Q(S^i_t, a^i_t) \leftarrow (1 - \alpha) Q(S^i_t, a^i_t) + \alpha \left(r^i_t + \gamma \max_{a^j} Q(S^j_{t+1}, a^j)\right),
\end{equation*}
%
\noindent where $\alpha$ is the learning rate, $\gamma$ is the discount factor,
and $S_{t+1}^j$ is the next state observed at the receiving node $\mathfrak{n}_j$.

\noindent \textbf{Training and Loss Function:}
To handle function approximation and improve training stability, we adopt a \gls{ddqn} approach~\cite{van2016deep}. The \gls{td} error is minimized via the following loss:

\begin{equation} \label{eq:loss}
    L(\theta) = \mathbb{E}\left[ \left( r^i_t + \gamma \overbrace{\max_{a^j} Q(S_{t+1}^{j}, a^{j}; \theta^-)}^{\text{Target Network}} - \overbrace{Q(S^i_t, a^i_t; \theta)}^{\text{Q-Network}} \right)^2 \right],
\end{equation}

\noindent where $Q(\cdot;\theta)$ is the online Q-network, and $Q(\cdot;\theta^-)$ the target network, updated periodically to stabilize training. Parameters $\theta$ are optimized with \gls{sgd} over batches sampled from the experience buffer $D_g$.




 \begin{algorithm}[h]
 \caption{\gls{gatmarl}: Exploration-to-Exploitation}
 \begin{algorithmic}[1]
 \renewcommand{\algorithmicrequire}{\textbf{Initialize:}}
 \REQUIRE Import $Q(s,a;\theta)$ and target $Q(s,a;\theta^-)$
 \REQUIRE Set exponential decay for $\varepsilon_t$
    \FOR {episode $t=1,2,\dotsc,\tau$}
        \STATE Update network edges $\mathcal{E}$
        \FOR{rover $\mathfrak{r}_i \in \mathfrak{R}$}
            \STATE Generate $N$ information packets
            \STATE Observe local state $S^i_t$
            \IF{$u \sim U(0,1) < \varepsilon_t$}
                \STATE Select random action $a^i_t$
            \ELSE
                \STATE Select $a^i_t = \arg\max_{a} Q(S^i_t, a;\theta)$
            \ENDIF
            \STATE Forward $\min(R(i,j), |\mathcal{B}_i|)$ packets to $\mathfrak{r}_j$ selected by $a^i_t$
            \STATE Compute reward $r^i_t$ using Eq.~\eqref{eq:reward}
            \STATE Observe next state $S^j_{t+1}$ from $\mathfrak{r}_j$
            \STATE Store transition $(S^i_t, a^i_t, r^i_t, S^j_{t+1})$ in $D_g$
            \STATE Perform SGD step using DDQN loss (Eq.~\eqref{eq:loss})
        \ENDFOR
    \ENDFOR 
 \end{algorithmic}
 \label{alg:montecarlo} 
\end{algorithm}

\begin{figure*}[t]
    \centering
    \begin{subfigure}{0.48\textwidth}
        \centering
        \includegraphics[width=\linewidth]{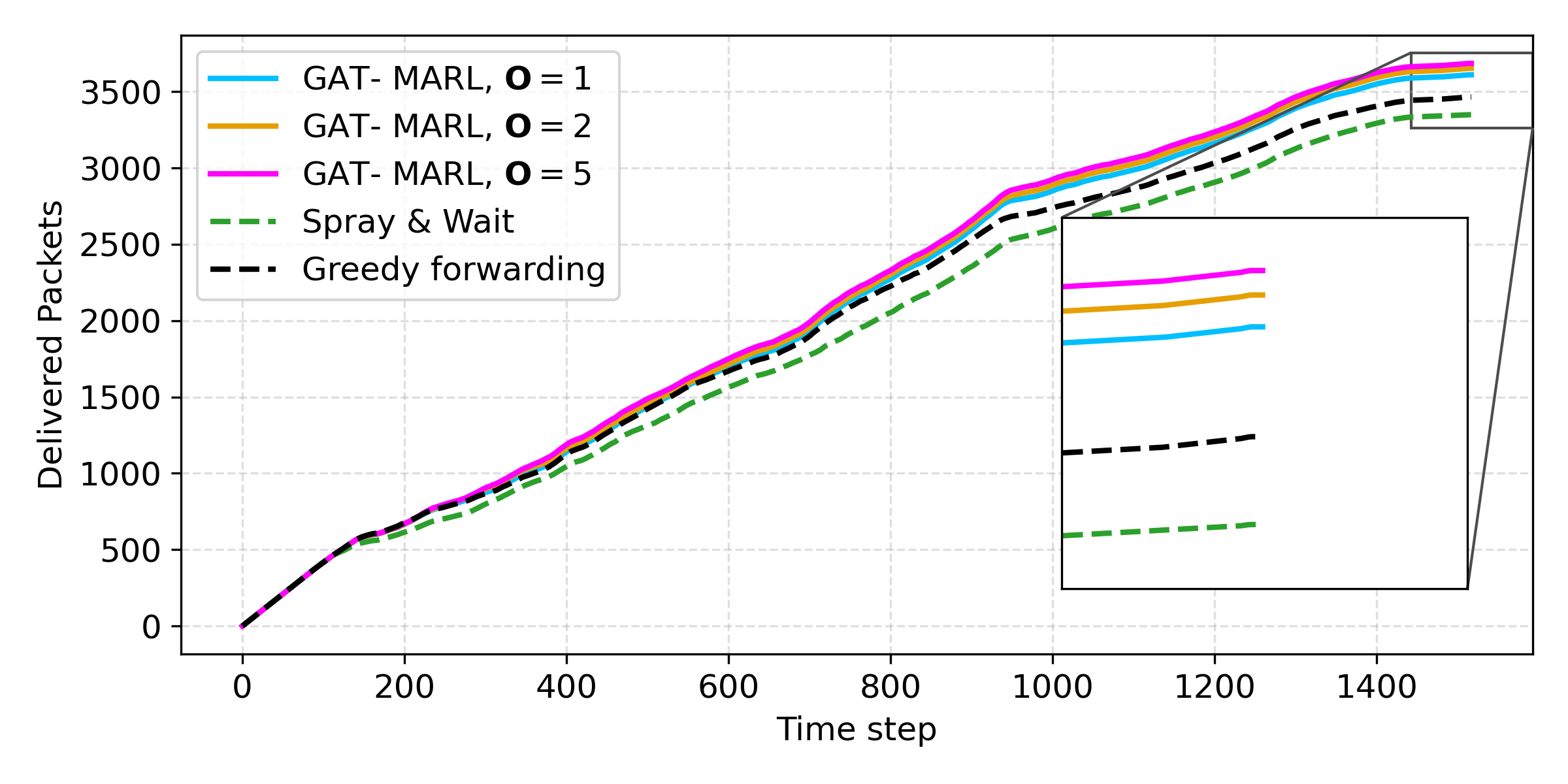}
        \caption{Delivered unique packets.}
        \label{fig:delivered}
    \end{subfigure}%
    \hfill
    \begin{subfigure}{0.48\textwidth}
        \centering
        \includegraphics[width=\linewidth]{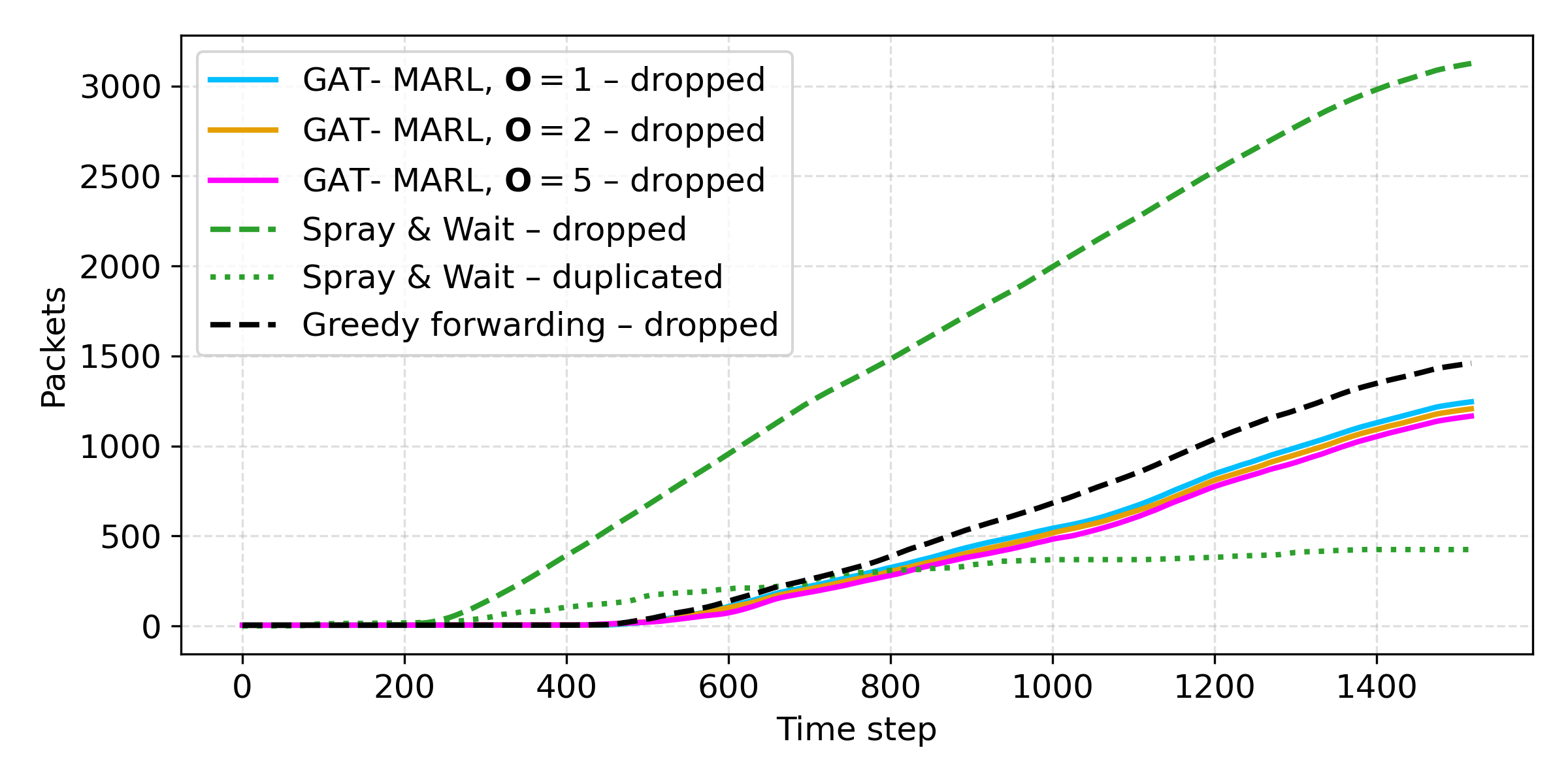}
        \caption{Duplicated delivered and dropped packets.}
        \label{fig:losses}
    \end{subfigure}

    \begin{subfigure}{0.48\textwidth}
        \centering
        \includegraphics[width=\linewidth]{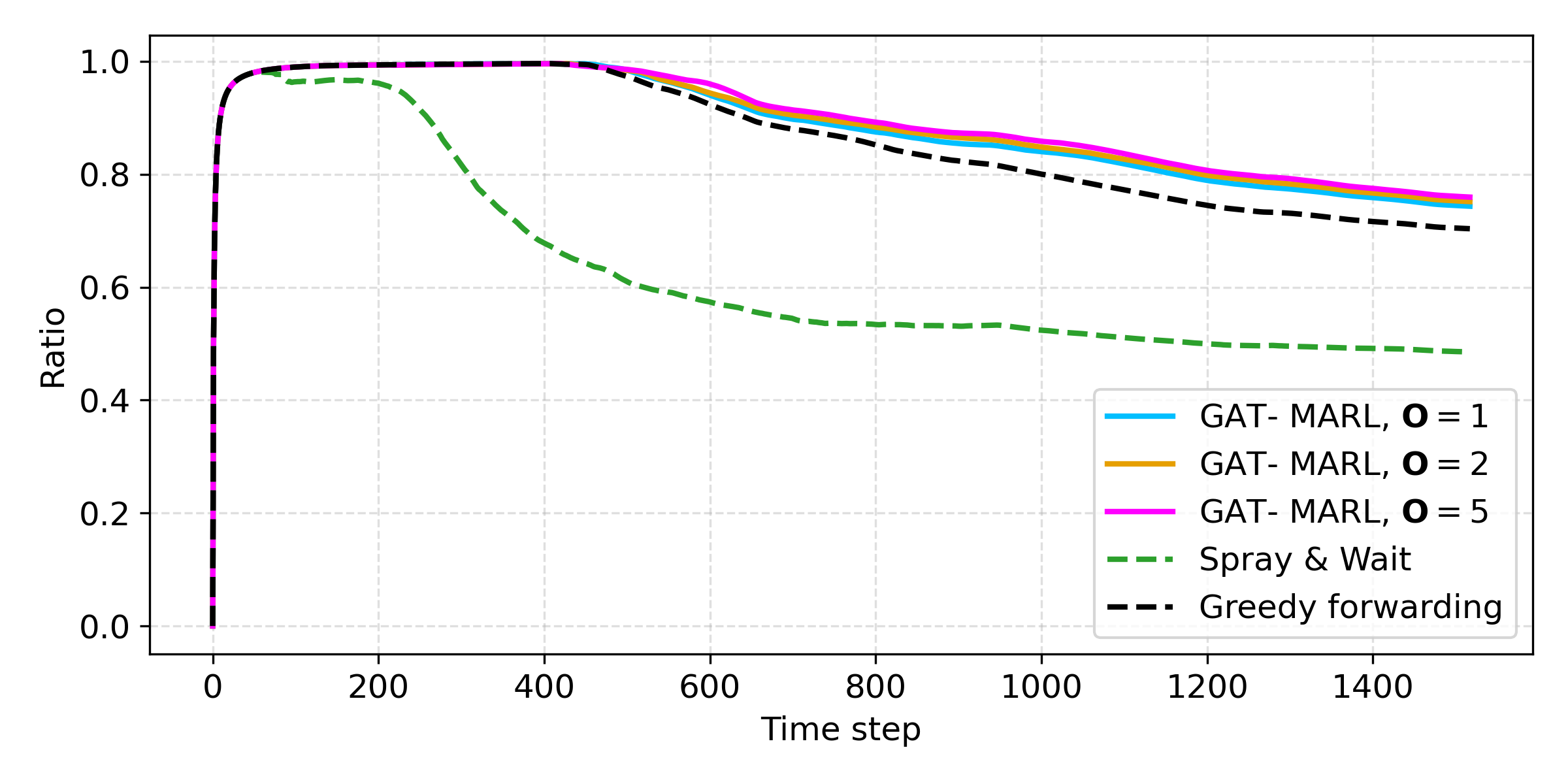}
        \caption{Packet delivery ratio including duplicated packets.}
        \label{fig:ratio}
    \end{subfigure}%
    \hfill
    \begin{subfigure}{0.48\textwidth}
        \centering
        \includegraphics[width=\linewidth]{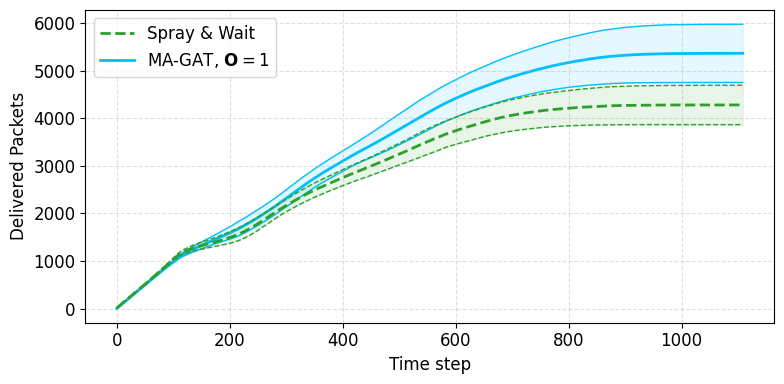}
        \caption{10 rovers in 150 unseen scenarios with $\pm1\sigma$ standard deviation.}
        \label{fig:10_rovers}
    \end{subfigure}
    
    \caption{Monte Carlo comparison of \emph{Spray and Wait}, \emph{Greedy forwarding}, and \gls{gatmarl} under different objective horizons $\mathbf{O}$. \ref{fig:10_rovers} shows generalization to exploration episodes $\tau$ with 10 rovers when the model was only trained on 3–5 rovers. 
    }
    \label{fig:results}
    \vspace{-0.5cm}
\end{figure*}

\section{Experimental Setup and Evaluation} \label{sec:results}

\noindent We evaluate our approach using a custom packet-level simulator developed in Python. The simulator supports flexible modeling of lunar exploration scenarios, including obstacle-aware mobility, intermittent communication links, and data packet routing through a \gls{ldtn}. 
Packets $p \in \mathcal{P}$ are implemented as explicit object instances, each encapsulating metadata such as the origin rover and the specific region of $\mathcal{M}_p$ the data pertains to. 
We evaluated the proposed \gls{gatmarl} policy, as well as two adapted classical non-learning strategies (\emph{Greedy forwarding} and \emph{Spray and Wait}). The environment setup and system components are fully aligned with the model described in Sec.~\ref{sec:systemmodel}. A typical outcome of one exploration episode of duration $\tau$ is illustrated in Fig.~\ref{fig:Exploration}. 

\noindent\textbf{Spray and Wait}~\cite{spyropoulos2005spray}\textbf{:} This controlled flooding algorithm operates in two phases. 
In (1) the \emph{Spray phase}, when a rover $\mathfrak{r}_i$ generates a new packet $p_i$, it initializes the packet’s copy counter {(which represents the maximum number of copies of the packet that will be created in the network)} as $L_{p_i} = \sqrt{|\mathfrak{R}|}$. 
When $\mathfrak{r}_i$ establishes a link $e_{ij}$ with another rover $\mathfrak{r}_j$, it generates and forwards a copy of the packet to $\mathfrak{r}_j$ if $L_{p_i} > 1$. The received copy $p_j$ is assigned 
$L_{p_j} = \left\lfloor \frac{L_{p_i}}{2} \right\rfloor$,
and {the sender} $\mathfrak{r}_i$ updates its packet's counter to $L_{p_i} = L_{p_i} - L_{p_j}$. 
Packets with $L > 1$ are considered \emph{active} and can continue to be spread. Once $L = 1$, the packet becomes \emph{inactive}, and (2) the \emph{Wait phase} starts, where nodes hold the packet until they directly encounter the destination. 
{We note} that $\mathfrak{r}_i$ will not attempt to forward any packets 
to $\mathfrak{r}_j$ if it is at capacity ($|\mathcal{B}_j| = B$). 
The \gls{fifo} policy is particularly valuable in this context, as it serves a dual purpose: (i) {older} packets are more likely to have distributed copies already, increasing delivery chances and (ii) inactive packets can only be forwarded directly to the destination,
making them difficult to offload from the buffer in the absence of direct contact with the lander—a common case in \glspl{ldtn} when the assigned area $\mathcal{M}_{\mathfrak{r}_i}$ is far from $\mathcal{L}$. 
The main advantage of this algorithm is that it operates without requiring global knowledge of the network topology or link availability—each node makes forwarding decisions based solely on its one-hop neighbors. 
We note that rovers avoid sending packets to neighbors that already hold a copy, assuming this knowledge is free—potentially overestimating performance.

\noindent\textbf{Greedy forwarding:} 
This method is an adaptation of the shortest path algorithm~\cite{dijkstra1959} for operation over a \gls{ldtn}. When a rover $\mathfrak{r}_i$ generates a packet $p_i$, it attempts to compute a shortest path to the lander $\mathcal{L}$ using the current network snapshot, i.e., the locally observed subgraph of the current connectivity state. If $\mathcal{L}$ is reachable from $\mathfrak{r}_i$ in that snapshot, $p_i$ is forwarded to the next hop $\mathfrak{r}_j$ along the path. Otherwise, $p_i$ is temporarily stored in $\mathfrak{r}_i$’s buffer $\mathcal{B}_i$. The main limitation of this approach lies in its dependence on up-to-date link state information, which requires frequent flooding of the network with control messages upon any change in topology. 
This overhead results in higher resource consumption and limits the scalability of the algorithm in highly dynamic networks. 
We note that this overhead is not modeled in our simulator: \emph{Greedy forwarding} assumes access to up-to-date topology at no cost, which overestimates its performance.

\noindent \textbf{\gls{gatmarl}:} 
{The \gls{gatmarl} model consists of a lightweight 2-layer} 
\gls{gat} followed by a fully connected prediction head. The first \gls{gat} layer uses 8 attention heads projecting input features from $32 \times 64$ to $32 \times 512$, followed by a second \gls{gat} layer reducing it to $32 \times 64$. A two-layer MLP head further processes the output with dimensions $64 \rightarrow 32 \rightarrow A$, where $A$ is the number of possible actions. Dropout (20$\%$) is applied after each layer to improve generalization. The model takes as input the feature matrix $\mathbf{X}_\mathfrak{r}$ and adjacency matrix $\mathbf{A}_\mathfrak{r}'$ (Fig.~\ref{fig:LDTN}), and outputs one Q-value per available next-hop action. The final trained model is compact, occupying only 188KB when serialized using the \verb|PyTorch| library~\texttt{.pt} file format, requiring minimal onboard computation and energy consumption. 


\noindent A Monte Carlo evaluation is performed across $\tau=30$ exploration episodes (as shown in Fig.~\ref{fig:Exploration}) with different maps $\mathcal{M_\tau}$ formed by randomly generated obstacle layouts and {a variable number of rovers between 3 and 5} 
per $\tau$. 
The evaluation follows a three-phase curriculum: (1) experience collection via full random actions for 10,000 steps, which are then used to train the \gls{gatmarl} model; (2) gradual shift from exploration to exploitation at each $\tau$, where the model is trained at every step (Alg.~\ref{alg:montecarlo}); and (3) pure exploitation under frozen policy: no training or experience collection is performed here. {Each phase comprises 10 $\tau$ and each $\tau$ consists of approximately 1000–1450 time steps $t$, depending on the number of rovers and the obstacle distribution in $\mathcal{M}_\tau$.} 

\noindent Figs.~\ref{fig:delivered},~\ref{fig:losses},~\ref{fig:ratio} present the average results per $\tau$ from the final phase, where the trained model is evaluated in unseen scenarios without further learning. We compare \emph{Spray and Wait}, \emph{Greedy forwarding}, and \gls{gatmarl}. 
For the latter, we evaluate different values of the objective horizon $\mathbf{O}$, defined as the number of upcoming objectives $\mathbf{o}$ considered by each rover when computing its \gls{ttl}.
\gls{gatmarl}  
{achieves  higher delivery rates}
(Fig.~\ref{fig:delivered})—with $\mathbf{O}=1$, it delivers $4.2\%$ and $7.8\%$ more packets than \emph{Greedy} and \emph{Spray and Wait}, respectively, and with $\mathbf{O}=5$, the gains increase to $6.4\%$ and $10\%$, respectively.
Critically, it also achieves fewer drops and no duplications (Fig.~\ref{fig:losses}), where \emph{Spray and Wait} delivers up to $424$ duplicated packets per $\tau$ on average. Finally, the proposed approach results in  better delivery ratios (Fig.~\ref{fig:ratio}), i.e., the number of unique packets delivered per packet created, with values of $74.35\%$ for \gls{gatmarl} ($\mathbf{O}=1$), $75.96\%$ for \gls{gatmarl} ($\mathbf{O}=5$), compared to $70.36\%$ for \emph{Greedy} and $48.54\%$ for \emph{Spray and Wait}. 
The performance gap widens for \emph{Spray and Wait} due to inefficiencies from excessive packet duplication. 
Overall, \gls{gatmarl} benefits from larger $\mathbf{O}$ values, effectively leveraging short-term mobility forecasts to improve routing.

\noindent To evaluate scalability and generalization, we test the same \gls{gatmarl} model—trained on only 20 $\tau$ with 3–5 rovers—on 150 new $\tau$ involving 10 rovers, without any additional training. As shown in Fig.~\ref{fig:10_rovers}, \gls{gatmarl} clearly outperforms \emph{Spray and Wait},
which delivers on average 4278 unique packets out of 5498, corresponding to a delivery rate of $77.8\%$; in contrast, \gls{gatmarl} delivers 5362 unique packets, achieving $97.5\%$. This gap arises because \emph{Spray and Wait} rapidly congests the network by creating multiple copies per packet, while \gls{gatmarl} learns to manage congestion more effectively, keeping the network load under control, also when it is not congested. We exclude \emph{Greedy forwarding} from the figure for two main reasons: (1) its simulated performance is similar to \gls{gatmarl} in these scenarios; but (2) it relies on up-to-date global topology information, requiring frequent flooding of control messages—a highly unrealistic assumption in dynamic scenarios with 10 agents, where the topology changes an average of 156 times per $\tau$. 

\noindent Overall, these results show that \gls{gatmarl} holds significant promise for achieving high delivery rates without duplication or the overhead of global topology discovery, potentially enabling bolder, beyond-line-of-sight exploration for future autonomous robotic missions.

\section{Conclusions} \label{sec:conclusions}

\noindent We presented a decentralized learning-based routing framework for multi-robot exploration in \glspl{ldtn}. By leveraging \glspl{gat} within a multi-agent \gls{rl} paradigm, our method enables each rover to make efficient, fully local routing decisions based on partial observations—without requiring global topology updates or packet duplication.
The proposed \gls{gatmarl} policy demonstrates strong performance across a wide range of simulated scenarios, consistently outperforming classical \gls{dtn} baselines in terms of delivery rate, redundancy, and loss. Moreover, it generalizes effectively to larger teams without retraining, indicating its potential for scalability in future mission deployments.
These results show that decentralized, attention-based policies can enable robust and scalable communication strategies in autonomous distributed planetary networks, supporting future science-driven missions.

\bibliographystyle{IEEEtran}
\bibliography{main}
\end{document}